\pdfoutput=1

\documentclass[11pt]{article}

\usepackage[final]{acl}

\usepackage{times}
\usepackage{latexsym}

\usepackage[T1]{fontenc}

\usepackage[utf8]{inputenc}

\usepackage{microtype}

\usepackage{inconsolata}

\usepackage{graphicx}

\usepackage[linesnumbered,ruled,vlined]{algorithm2e}
\usepackage{pifont}
\usepackage{booktabs}
\usepackage{makecell}
\usepackage{multirow}
\usepackage{subcaption}
\usepackage{nolbreaks}
\SetAlgoSkip{}

\title{\textsc{Seg2Act}: Global Context-aware Action Generation \\ for Document Logical Structuring}

\author{
  Zichao Li${}^{1,2,}$\thanks{Equal contribution.},
  Shaojie He${}^{1,2,}$\footnotemark[1],
  Meng Liao${}^{3,}$\thanks{Corresponding author.},
  Xuanang Chen${}^{1}$, \\
  \textbf{Yaojie Lu}${}^{1,}$\footnotemark[2], 
  \textbf{Hongyu Lin}${}^{1}$,
  \textbf{Yanxiong Lu}${}^{3}$,
  \textbf{Xianpei Han}${}^{1}$,
  \textbf{Le Sun}${}^{1}$
  \\
  ${}^{1}$Chinese Information Processing Laboratory, Institute of Software, \\
  Chinese Academy of Sciences, Beijing, China\\
  ${}^{2}$University of Chinese Academy of Sciences, Beijing, China \\
    ${}^{3}$Search Team, WeChat, Tencent Inc., China \\ 
 {\tt \{lizichao2022,heshaojie2020,chenxuanang,luyaojie\}@iscas.ac.cn} \\
   {\tt \{hongyu,xianpei,sunle\}@iscas.ac.cn \{maricoliao, alanlu\}@tencent.com} \\
}

\begin{document}
\maketitle
\begin{abstract}
Document logical structuring aims to extract the underlying hierarchical structure of documents, which is crucial for document intelligence.
Traditional approaches often fall short in handling the complexity and the variability of lengthy documents.
To address these issues, we introduce \textsc{Seg2Act}, an end-to-end, generation-based method for document logical structuring, revisiting logical structure extraction as an action generation task.
Specifically, given the text segments of a document, \textsc{Seg2Act} iteratively generates the action sequence via a global context-aware generative model, and simultaneously updates its global context 
 and current logical structure based on the generated actions. 
Experiments on ChCatExt and HierDoc datasets demonstrate the superior performance of  \textsc{Seg2Act} in both supervised and transfer learning settings\footnote{The publicly available code are accessible at \url{https://github.com/cascip/seg2act}.}.

\end{abstract}

\section{Introduction}

\begin{figure}[!th]
    \centering
    \scalebox{0.47}{
    \includegraphics{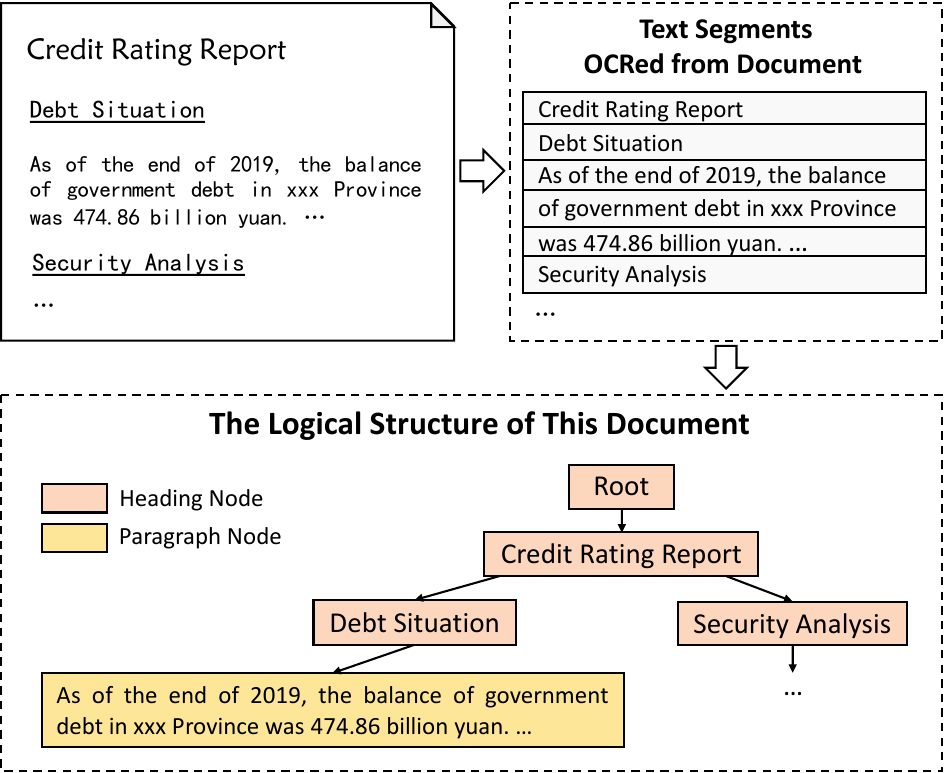} 
    }
    \caption{The illustration of document logical structuring task, which aims to transform text segments into a hierarchical tree structure containing the document's headings and paragraphs.}
    \label{figures.task_definition}
\end{figure}

Document logical structuring is an essential task for document understanding, which aims to extract the underlying logical structure of documents \cite{DBLP:conf/icpr/TsujimotoA90,DBLP:phd/us/Summers98,DBLP:conf/drr/MaoRK03,DBLP:journals/ijdls/LuongNK10,DBLP:journals/nle/PembeG15,DBLP:conf/emnlp/GopinathWS18,maarouf-etal-2021-financial}. 
As shown in Figure \ref{figures.task_definition}, document logical structuring transforms a document into a hierarchical logical tree composing of headings and paragraphs.
Understanding a document's logical structure will benefit numerous downstream tasks, such as information retrieval \cite{DBLP:conf/emnlp/LiuHZYXY21}, abstractive summarization \cite{DBLP:conf/emnlp/QiuC22}, and assisting large language models in question answering over long structured documents \cite{DBLP:journals/corr/abs-2309-08872}.

\begin{figure*}[!t]
    \centering
    \scalebox{0.5}{
    \includegraphics{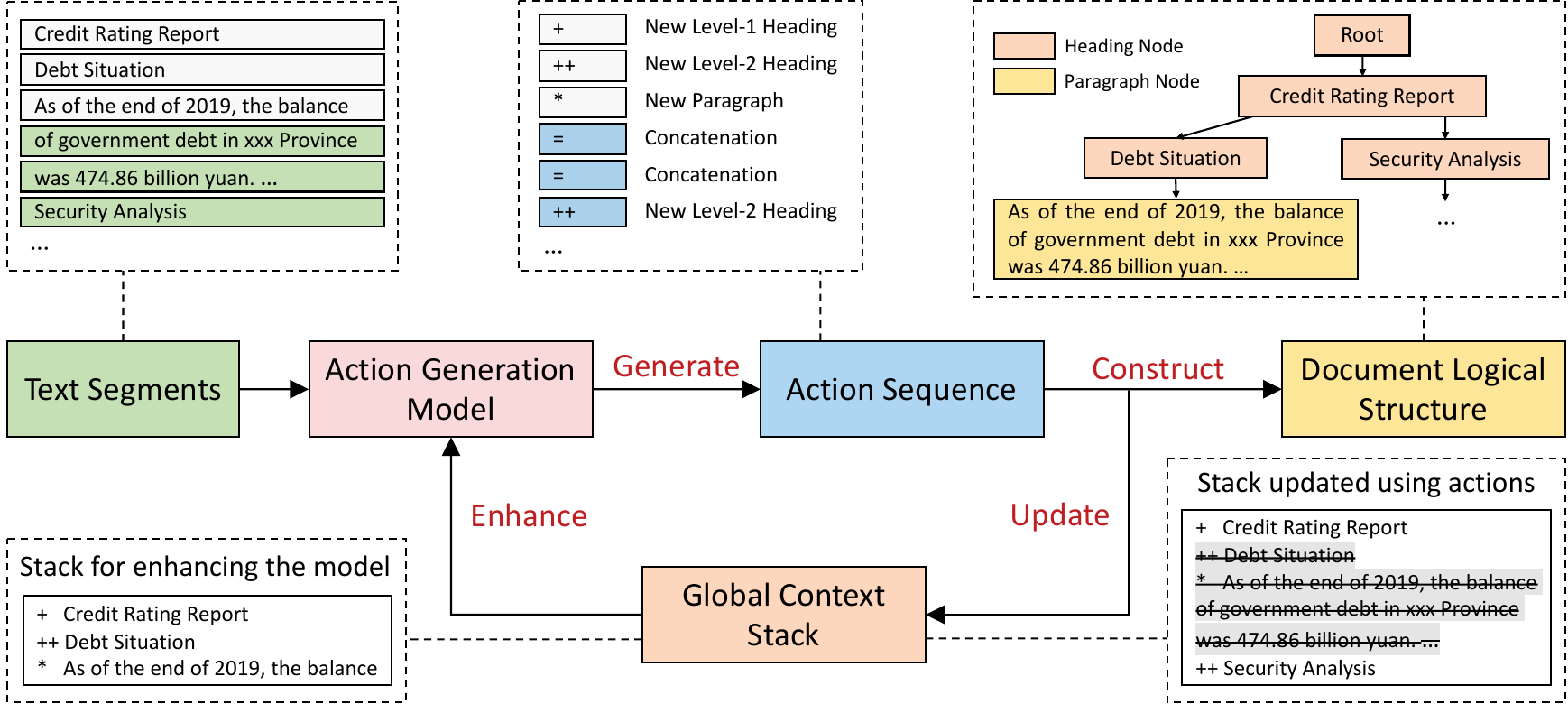} 
    }
    \caption{ 
    A generation step of \textsc{Seg2Act}. The action generation model converts current text segments into actions to incrementally construct the document logical structure. A global context stack is maintained to enhance the model's global awareness, while the generated actions then being employed to update the stack.
}
    \label{figures.framework}
\end{figure*}

Document logical structuring is challenging due to the complexity of text segment dependencies in documents and the diversity of logical structures across various documents.
Firstly, real-world documents are mostly multi-page, lengthy and with complex structures, while OCR tools often break content into short and incomplete lines rather than complete paragraphs. Such inconsistency between text segments and hierarchical structure poses a significant challenge to tracking and formulating text semantics and long-range dependencies.
Secondly, due to the diversity of logical structures in various documents (e.g., financial report and scientific literature), it is very difficult to design a unified approach with strong generalization abilities, i.e., it can solve different types of documents.

Currently, most document logical structuring methods first decompose the extraction of logical structure into multiple separated subtasks (mostly including feature extraction, heading detection and nodes relationship prediction),  then compose the components of different subtasks in a pipeline to predict the final document logical structure \cite{DBLP:conf/bdc/RahmanF17,DBLP:conf/icdar/BentabetJF19,DBLP:conf/icpr/HuZZDW22,DBLP:conf/emnlp/Wang0023}.
The main drawbacks of these methods are: 
1) By encoding fragmented text segments independently, these methods cannot capture the global information of documents and often result in semantic loss. 
2) By pairwise predicting the relationship between text segments, these methods often ignore the long-range dependencies and result in sub-optimal structures.
3) The pipeline framework suffers from the error propagation problem. 
Due to the varieties of document structures, it is very challenging to design the optimal composition architecture manually for different types of documents.

To address these issues, in this paper, we propose \textsc{Seg2Act}, a global context-aware action generation approach for document logical structuring. 
As illustrated in Figure \ref{figures.framework}, instead of decomposing the extraction of logical structure into subtasks, we revisit structure extraction as an action generation task. 
Specifically, sequentially feeding a document's text segments, a global context-aware generative model is employed to generate a sequence of actions for document logical structuring. 
We propose three types of actions, each corresponding to an operation that maps text segments to the logical structure, applicable across various types of documents.
Furthermore, during the structuring process, \textsc{Seg2Act} maintains a global context stack which selectively stores crucial parts of global document information, expressing long-range dependencies in a concentrated manner. 
In this way, \textsc{Seg2Act} can effectively handle various document types, generate the logical structure of a document in an end-to-end manner, and leverage global document information for text segment encoding and structure generation.
Experiments on ChCatExt and HierDoc datasets demonstrate that \textsc{Seg2Act} achieves superior performance in both supervised and transfer settings, verifying the effectiveness and the generalization ability of the proposed method.



Our contributions are summarized as follows:
1) This is the first work to make the logical structure extraction as an one-pass action generation task, which is more generalizable and easy to implement.
2) A generation framework called \textsc{Seg2Act} is proposed, which adopts a global context-aware generative model to better encode the semantics of text segments and model the long-range dependencies between them.
3) \textsc{Seg2Act} significantly outperforms baselines in both supervised and transfer settings, showing its effectiveness and the generalization ability.






\section{\textsc{Seg2Act}: Document Logical Structuring as Action Generation}



\subsection{Overview}
As mentioned, this work considers document logical structuring task as an action generation task.
That is, given a sequence of text segments $X=x_1, ..., x_N$, the goal is to produce a sequence of actions $Y=y_1, ..., y_N$, which are further used to construct the logical structure $T$ of the document. 
The overall framework of \textsc{Seg2Act} is depicted in Figure \ref{figures.framework}. 
Specifically, given a sequence of text segments, a window with $w_{I}$ segments is input to the action generation model iteratively, to obtain an action sequence consisting of three types of actions.
During a generation step, the previous actions and segments are constructed as a global context stack, which can provide global information for the action generation model. 
After that, the generated actions update both document logical structure and global context stack simultaneously.
Once all text segments have been processed, the complete logical structure of target document will be produced.


\subsection{Actions for Document Logical Structuring}\label{sec.actions}
The logical structure is a hierarchical tree composed of heading and paragraph nodes, where the depth of a node represents its level. Before structuring, a level-0 heading node with no textual content is added as the root. To achieve one-pass structuring, we define three actions to map text segments to the logical structure:

\begin{itemize}
    \setlength{\itemsep}{1pt}
    \item \textbf{New Level-$k$ Heading}: this action signifies adding the text segment as a new level-$k$ heading node to the current document logical structure, with the last added level-$(k-1)$ heading node serving as its parent. We use $k$ consecutive ``+'' to represent it.
    \item \textbf{New Paragraph}: this action denotes adding the current segment as a new paragraph node to the document logical structure,  with the last added heading node serving as its parent. We use an asterisk ``*'' to represent it.
    \item \textbf{Concatenation}: this action indicates that the corresponding segment is an extension of the preceding text. It appends the text of the corresponding segment to the last added node of the current document logical structure. We use an equal sign ``='' to represent it.
\end{itemize}




Previous works, such as TRACER~\cite{DBLP:conf/icdar/ZhuZLYRWWHCC23}, also define a series of actions, but they are performed under pairwise local transitions, and a segment may participate multiple times due to the shift-reduce operation.
In contrast, \textsc{Seg2Act} establishes a one-to-one relationship between segments and actions, 
directly mapping text segments to specific positions in the document's logical structure. 
This design reduces the number of necessary predictions, resulting in a more efficient process.
\subsection{Action Generation Model}
The action generation model refers to a generative language model, which is adopted to convert text segments into action sequence by considering the global information.
Specifically, as illustrated in Table \ref{tables.global_structure_schema}, this action generation model takes a global context stack and the current input text segments as input to predict actions for constructing the logical structure.
In this section, we first describe the global context stack, which enhances the action generation as it provides global information. Then, we present the multi-segment multi-action strategy, wherein $w_{I}$ segments are converted into $w_{O}$ actions at each step, which broadens the model's perspective and accelerates the construction process.


\begin{table}[t]
\centering
\scalebox{0.7}{
    \begin{tabular}{l p{8.5cm}}
    \toprule
    \multicolumn{2}{l}{\#\#\# STACK:} \\
    + & Government Bonds Credit Rating Report \textcolor{red}{$\hookleftarrow$} \\
    ++ & Credit Quality Analysis for this Series \textcolor{red}{$\hookleftarrow$} \\
    +++ & Use of Proceeds \textcolor{red}{$\hookleftarrow$} \\
    $*$ & The funds raised from the Government Bonds are ... and projects related to agriculture, \textcolor{red}{$\hookleftarrow$} \\\\
    \multicolumn{2}{l}{\#\#\# SEGMENT:} \\
    ~ & forestry, water resources and social services. \textcolor{red}{$\hookleftarrow$} \\
    ~ & Payment Security Analysis \textcolor{red}{$\hookleftarrow$} \\
    ~ & The proceeds for the projects funded by this bond issue are derived from project operational revenues \textcolor{red}{$\hookleftarrow$} \\\\
    \multicolumn{2}{l}{\#\#\# ACTION:} \\
    \multicolumn{2}{l}{= \textcolor{red}{$\hookleftarrow$}~~~~+++ \textcolor{red}{$\hookleftarrow$}~~~~ * \textcolor{red}{$\hookleftarrow$}}\\
    \bottomrule
    \end{tabular}
}
\caption{A demonstration example of the model template in a single prediction step. It utilizes the global context stack and multi-segment multi-action strategy. ``\textcolor{red}{$\hookleftarrow$}'' denotes a line break.
}\label{tables.global_structure_schema}
\end{table}


\subsubsection{Global Context Stack} \label{sec.gcs}
To keep the action generation model informed about the ongoing construction process, we design a global context stack to provide global information to aid the model in decision-making.
Specifically, as shown in Figure~\ref{figures.framework} and Table~\ref{tables.global_structure_schema}, we utilize the same symbols (``+'' and ``*'') as actions introduced in Section~\ref{sec.actions} to organize previous text.


The global context stack selectively contains a subset of nodes from the constructed logical structure.
Initially, the stack contains only the root node. 
For each generation step, it is updated according to the generated actions:
\textbf{New Level-$k$ Heading} continuously pops nodes until the stack top is a level-$(k-1)$ heading, then pushes the new level-$k$ heading node. 
\textbf{New-Paragraph} pops the paragraph node (if any) from the top of the stack, then pushes the new paragraph node.
\textbf{Concatenation} appends the current text segment to the top node of the stack.
Thus, the stack stores the last added node at the top, followed by all the nodes along the upward backtracking path in the hierarchical tree,
which we intuitively regard as being closely related to the current structuring.

Based on this approach, the global context stack models the long-distance dependencies in a centralized manner, enabling global information to be facilitated within a limited input length.


\subsubsection{Multi-segment Multi-action Strategy}
Since documents are segmented at the line level, there would be a lot of text segments for a document waiting for action prediction. For example, the average number of text segments in the HierDoc dataset \cite{DBLP:conf/icpr/HuZZDW22} is $853.38$.
Therefore, if we process text segments one by one, it is not only insufficient for capturing the complete semantics, but also inefficient for obtaining all actions of segments.
To this end, we propose a multi-segment multi-action strategy to strengthen our \textsc{Seg2Act} framework to be more practical.

Specifically, we not only extend the length of the input segment window, denoted as $w_{I}$, but also extend the output action window's length, denoted as $w_{O}$.
On one hand, in a single prediction step, the action generation model receives $w_{I}$ consecutive text segments, which allows the input segment window to encompass a more extensive range of contextual information, facilitating informed decision-making by the model.
On the other hand, we can instruct the action generation model to predict $w_O$ actions in a single step to speed up the whole generation process, thereby reducing the required number of prediction steps to $\lceil N/w_O \rceil$, where $1 \leq w_O \leq w_I$. 
When $w_I=w_O$, it is our default setting, representing the one-pass mode.






\subsection{Model Training and Inference}

In this section, we first describe how to train the action generation model, and then introduce the inference process that includes constraints. 

\subsubsection{Training}


The training dataset consists of a collection of documents, each denoted as $D$.
Each document is comprised of text segments $X=x_1, ..., x_N$, along with a corresponding sequence of action annotations $Y=y_1, ..., y_N$.
More details of data pre-processing can be found in Appendix \ref{data_preprocessing}.
We optimize the global context-aware action generation model using teacher-forcing cross-entropy loss, which is defined as:
\begin{equation}
    \mathcal{L} = -\sum_{i=1}^{|D|} \log P(y_{i:i+w_{I}-1} |s_{i}, x_{i:i+w_{I}-1};\Theta)
\end{equation}
where $s_i$ represents the global context stack associated with the text segment $x_{i}$ and $\Theta$ denotes the parameters of the model.
For multi-segment multi-action strategy, $w_{I}$ represents the input segment window size. 
The model learns to predict actions aligned with $w_{I}$, namely, $w_I=w_O$, which means the number of predicted actions is equal to the number of input segments during model training.

\subsubsection{Inference} \label{inference}
Given a sequence of text segments from a document, as shown in Algorithm \ref{algorithm}, we utilize the trained action generation model to generate actions for segments, and then parse the actions to obtain the logical structure.
During inference, after setting the input size of segments $w_{I}$, we can use $w_{O}$ to control the speed of the iterative action execution process.
The greedy search algorithm is used to generate the action sequence.
At each generation step, we parse $w_O$ actions to update the document logical structure, as outlined in Section \ref{sec.actions}, and update the global context stack as described in Section \ref{sec.gcs}. 
After all segments are processed, we can obtain the final logical structure for the document.

\begin{algorithm}[!t]
\small
\SetAlgoLined
\SetKwInOut{Input}{Input}
\SetKwInOut{Output}{Output}
\SetKwInOut{Initialize}{Initialize}
\Input{Text segments $X=x_1,...,x_N$, \\input segment window's length $w_I$, \\output action window's length $w_O$.}
\Output{Document logical tree structure $T$.}
\BlankLine
\Initialize{root $\gets$ HeadingNode(), \\stack $S \gets$ [root], tree $T \gets$ [root].}
\For{$i \gets 1$ \KwTo $\lceil N/w_O \rceil$}{
  segments $\gets [x_{(i - 1) \cdot w_O + 1}, ..., x_{(i - 1) \cdot w_O + w_I}]$\;
  actions $\gets$ Model($S$, segments)\;
  \For{$j \gets 1$ \KwTo $w_O$}{
    \If{\textnormal{actions[$j$] = ``New Level-$k$ Heading''}}{
      node $\gets$ HeadingNode(segments[$j$])\;
      UpdateStackAndTree($S$, $T$, node)\;
    }
    \ElseIf{\textnormal{actions[$j$] = ``New Paragraph''}}{
      node $\gets$ ParagraphNode(segments[$j$])\;
      UpdateStackAndTree($S$, $T$, node)\;
    }
    \ElseIf{\textnormal{actions[$j$] = ``Concatenation''}}{
      ConcatText($S$, $T$, segments[$j$])\;
    }
  }
}
\Return $T$\;
\caption{\nolbreaks{Text segments to logical structure}} \label{algorithm}
\end{algorithm}

To ensure the validity of the generated action sequence and the effective updating of the logical structure and the stack, we apply some hard constraints. 
For example, tokens outside of a predefined set will be banned, and the concatenation action "=" cannot be generated when the stack contains only the root node. 
All constrains and the execution method can be found in Appendix \ref{constraints}.
In rare cases where the output number of actions mismatches $w_{I}$, we treat these as failures, skip these segments, and continue to the next generation step.

\section{Experiments}

This section evaluates \textsc{Seg2Act} by conducting experiments in both supervised learning and transfer learning settings. 

\subsection{Experimental Settings}

\paragraph{Datasets.}
We conduct experiments on the following datasets: 1) ChCatExt corpus \cite{DBLP:conf/icdar/ZhuZLYRWWHCC23}, which contains text segments from 650 Chinese documents and corresponding logical structures.
2) HierDoc corpus \cite{DBLP:conf/icpr/HuZZDW22}, consisting of 650 English scientific documents and corresponding Table-of-Content (ToC) structures, which contains only heading annotations.

\paragraph{Metrics.}
For evaluation, we use the same criteria in previous work, including F1-score and TEDS \cite{DBLP:conf/icpr/HuZZDW22,DBLP:conf/icdar/ZhuZLYRWWHCC23}. Additionally, we add a new criterion, DocAcc, to evaluate the accuracy of logical structures at the document level.

\textbf{DocAcc}. A prediction is considered to be correct only when the logical structure exactly matches the ground truth; otherwise, it is judged as incorrect.

\paragraph{Baselines.} 
We compare our method with the following two groups of baselines:

1) Baselines using text only: TRACER \cite{DBLP:conf/icdar/ZhuZLYRWWHCC23} is a transition-based framework for logical structure extraction, which predicts transition actions by encoding local pairwise text segments through a pre-trained language model. 

2) Baselines using text, layout and vision: MTD \cite{DBLP:conf/icpr/HuZZDW22} is a multi-modal method that utilizes pre-trained models to encode visual, textual, and positional document information, extracting ToC by attention and pairwise classification stages; 
CMM \cite{DBLP:conf/emnlp/Wang0023} is a three-stage framework that starts with a heuristic-based initial tree, then encodes nodes with pre-trained models, and finally refines the tree by moving or deleting nodes.

For our approach, we conduct the experiments of two settings:

1) \textsc{Seg2Act}. It is a global context-aware action generation method proposed in this paper, which generates the document logical structure in an end-to-end, one-pass manner.

2) \textsc{Seg2Act-T}. It is a modified version of TRACER, in which we utilize our proposed global context-aware generative model as the action parser, while still generating shift-reduce actions and following constraints akin to TRACER.

\begin{table}[t]
\centering
\scalebox{0.72}{
    \begin{tabular}{lccccc}
    \toprule
    \textbf{Method} & \textbf{Heading} & \textbf{Paragraph} & \textbf{Total} & \textbf{DocAcc} \\
    \midrule
    \multicolumn{5}{l}{\textit{\textbf{Methods using RBT3 as Backbone}}} \\
    TRACER & 90.49 & 84.33 & 82.39 & - \\
    TRACER$^{*}$ & 90.04 & 83.96 & 82.07 & 26.15 \\
    \midrule
    \multicolumn{5}{l}{\textit{\textbf{Methods using GPT2-Medium as Backbone}}} \\
    TRACER$^{*}$ & 91.15 & 88.53 & 85.40 & 47.38 \\
    \textsc{Seg2Act-T} (Ours) & 93.94 & 91.21 & 89.01 & 52.00 \\
    \textsc{Seg2Act} (Ours) & 94.88 & 92.99 & 90.96 & 57.23 \\
    \midrule
    \multicolumn{5}{l}{\textit{\textbf{Methods using Baichuan-7B as Backbone}}} \\
    TRACER$^{*}$ & 94.91 & 91.62 & 89.55 & 53.85 \\
    \textsc{Seg2Act-T} (Ours) & \textbf{96.01} & 93.98 & 92.39 & 58.46 \\
    \textsc{Seg2Act} (Ours) & \textbf{96.01} & \textbf{94.19} & \textbf{92.63} & \textbf{63.69} \\
    \bottomrule
    \end{tabular}
}
\caption{Overall performance on ChCatExt (Heading, Paragraph, Total nodes in F1-score and logical structure accuracy at the document level). TRACER$^{*}$ refers to our implemented results.}
\label{tables.main_result}
\end{table}

\begin{table}[t]
    \centering
    \scalebox{0.72}{
        \begin{tabular}{ccccc}
        \toprule
        \textbf{Method} & {\textbf{Modality}} & \textbf{Backbone} & {\textbf{HD}} & {\textbf{ToC}} \\ \midrule
        MTD & T+L+V & BERT+ResNet & 96.1 & 87.2 \\
        CMM & T+L & RoBERTa & 97.0 & 88.1 \\ \midrule
        \multirow{2}{*}{\textsc{Seg2Act} (Ours)} & \multirow{2}{*}{T} & GPT2-Medium & 96.3 & 93.3 \\
         &  &  Baichuan-7B & \textbf{98.1} & \textbf{96.3} \\
        \bottomrule
        \end{tabular}%
    }
    \caption{Heading detection (HD) in F1-score and ToC in TEDS (\%) of baselines and \textsc{Seg2Act} on HierDoc.}
    \label{tables.hierdoc_result}
\end{table}

\paragraph{Implementations.}
Our implementations are built upon Pytorch \cite{paszke2019pytorch}, Transformers \cite{wolf2020transformers}  and PEFT \cite{peft} libraries.
For both GPT2-Medium and Baichuan-7B backbone models \cite{radford2019language,baichuan2023}, we employ the AdamW \cite{DBLP:conf/iclr/LoshchilovH19} optimizer with a learning rate of $3 \times 10^{-4}$. The number of training epochs is set to 10, and the batch size is set to 128. We set the input segment window and output action window as $w_I=w_O=3$. 
Experiments are conducted on an NVIDIA A100 GPU. 
For the transfer learning experiments, we initially pre-train models on the Wiki corpus (provided by \citet{DBLP:conf/icdar/ZhuZLYRWWHCC23}) for 10,000 steps.
Besides, we utilize the LoRA \cite{DBLP:conf/iclr/HuSWALWWC22} technique to reduce the GPU memory overhead during Baichuan-7B training. We set the rank $r$ to 8 and the alpha value $\alpha$ to 16. All experiments are averaged results obtained from five different random seeds to ensure robustness and reliability.

\subsection{Results in Supervised Learning Setting}

\begin{table*}[!t]
\centering
\scalebox{0.78}{
    \begin{tabular}{c|c|c|cc|cc|cc|c|c|c}
    \toprule
    \multirow{3}{*}{\textbf{Test Set}} & \multirow{3}{*}{\textbf{Method}} & \multirow{3}{*}{\textbf{Zero-Shot}} & \multicolumn{6}{c|}{\textbf{Few-Shot}} & \multicolumn{3}{c}{\textbf{Full-Shot}} \\
    \cline{4-12}
    & & & \multicolumn{2}{c|}{\textbf{BidAnn}} & \multicolumn{2}{c|}{\textbf{FinAnn}} & \multicolumn{2}{c|}{\textbf{CreRat}} & \multirow{2}{*}{\textbf{BidAnn}} & \multirow{2}{*}{\textbf{FinAnn}} & \multirow{2}{*}{\textbf{CreRat}}\\
    & & & \textbf{3} & \textbf{5} & \textbf{3} & \textbf{5} & \textbf{3} & \textbf{5} & & & \\
    \midrule
    \multirow{3}{*}{\textbf{BidAnn}} & TRACER & 2.70 & 66.64 & 14.58 & 2.36 & 21.48 & 1.02 & 12.78 & 88.20 & 25.26 & 11.74\\
    & \textsc{Seg2Act-T} (Ours) & 42.92 & 96.53 & 97.07 & 86.45 & 85.89 & 88.78 & 89.53 & 99.40 & 95.74 & \textbf{73.49}\\
    & \textsc{Seg2Act} (Ours) & \textbf{56.25} & \textbf{99.31} & \textbf{99.45} & \textbf{90.30} & \textbf{96.89} & \textbf{95.59} & \textbf{97.12} & \textbf{99.72} & \textbf{98.07} & 69.92 \\
    \midrule
    \multirow{3}{*}{\textbf{FinAnn}} & TRACER & 11.39 & \textbf{67.04} & 15.51 & 3.52 & 32.57 & 1.77 & 25.29 & 8.10 & 68.59 & 20.17 \\
    & \textsc{Seg2Act-T} (Ours) & 28.98 & 26.17 & \textbf{28.18} & 42.37 & 58.21 & 47.71 & 48.15 & 32.47 & 76.87 & 46.04 \\
    & \textsc{Seg2Act} (Ours) & \textbf{43.30} & 25.00 & 23.51 & \textbf{56.17} & \textbf{75.19} & \textbf{48.38} & \textbf{55.11} & \textbf{47.92} & \textbf{85.17} & \textbf{60.25} \\
    \midrule
    \multirow{3}{*}{\textbf{CreRat}} & TRACER & 14.07 & \textbf{79.03} & 16.42 & 4.52 & 27.53 & 18.66 & 19.24 & 7.00 & 30.82 & 92.29 \\
    & \textsc{Seg2Act-T} (Ours) & 49.65 & 35.71 & 31.31 & \textbf{47.79} & 56.36 & 71.63 & 84.77 & 32.33 & 42.19 & 95.77 \\
    & \textsc{Seg2Act} (Ours) & \textbf{67.86} & 55.22 & \textbf{54.75} & 24.32 & \textbf{65.73} & \textbf{82.77} & \textbf{86.59} & \textbf{61.20} & \textbf{70.14} & \textbf{97.76} \\
    \bottomrule
    \end{tabular}
}
\caption{Performance (F1-score of total nodes) on transfer learning experiments in zero-shot, few-shot and full-shot settings on three sub-corpora of ChCatExt: 
bid announcements (BidAnn) with 100 documents, financial announcements (FinAnn) with 300 documents, and credit rating reports (CreRat) with 250 documents..}
\label{tables.shot_result}
\end{table*}

Table \ref{tables.main_result} shows the performance of text-only baselines on the ChCatExt, And Table \ref{tables.hierdoc_result} compares the performance of multi-modal baselines on HierDoc with text-only \textsc{Seg2Act}. We can see that:

1) \textbf{By generating the logical structure in an end-to-end manner, \textsc{Seg2Act} achieves state-of-the-art performance.} 
In Table \ref{tables.main_result}, \textsc{Seg2Act} predicts the document logical structure with high accuracy, and outperforms TRACER under the same Baichuan-7B backbone by +9.84 in DocAcc. Table \ref{tables.hierdoc_result} shows that our method performs better than multi-modal methods in both HD and TEDS, even though it only uses semantic information. These above indicate that our \textsc{Seg2Act} can better perceive the overall logical structures of the documents.

2) \textbf{Global contextual information plays a crucial role in document logical structuring.} In Table \ref{tables.main_result}, injecting global context stack into TRACER produces a general performance improvement. In both GPT2-Medium and Baichuan-7B backbones, \textsc{Seg2Act-T} surpasses TRACER in terms of F1-score for headings, paragraphs, total nodes and document-level accuracy. This highlights the significance of the global context stack for document logical structuring.

\subsection{Results in Transfer Learning Setting}

To assess the generalization of \textsc{Seg2Act}, we first pre-train the backbone model on the Wiki corpus and then conduct a series of transfer learning experiments under zero-shot, few-shot, and full-shot settings, as shown in Table \ref{tables.shot_result}.
For ease of presentation, we use the F1-score of total nodes as the representative metric. We observe that:

1) \textbf{The action generation framework of \textsc{Seg2Act} can learn general document structures instead of capturing type-specific features.} 
Compared with \textsc{Seg2Act-T}, \textsc{Seg2Act} attains average improvements of +10.65, +6.04, +15.28 for full-shot, few-shot, and zero-shot settings, exhibiting its superiority in various scenarios.

2) \textbf{\textsc{Seg2Act} can robustly resist data scarcity, displaying a quick adaptation capability.} Taking the case of 5-shot training as an example, \textsc{Seg2Act} only averages a slight drop of 3.98 compared to the full-shot setting.

\subsection{Ablation Study}
\label{section.ablation_study}

\subsubsection{Effects of Global Context Stack}

\begin{table}[t]
\centering
\scalebox{0.62}{
    \begin{tabular}{lcccc}
    \toprule
    \textbf{Method} & \textbf{Heading} & \textbf{Paragraph} & \textbf{Total} & \textbf{DocAcc} \\
    \midrule
    \textsc{Seg2Act} & \textbf{96.01} & 94.19 & 92.63 & \textbf{63.69} \\
    ~~~~- multi-segment multi-action & 95.49 & 93.32 & 91.56 & 62.15 \\
    ~~~~- GCS (symbol) & 95.71 & \textbf{94.28} & \textbf{92.69} & 57.23  \\
    ~~~~- GCS (text) & 90.92 & 87.52 & 83.85 & 50.77  \\
    ~~~~- GCS (both text and symbol) & 89.45 & 85.36 & 81.15 & 44.92  \\
    \bottomrule
    \end{tabular}
}
\caption{Performance on ChCatExt with ablated settings. GCS denotes the global context stack. }
\label{tables.ablation}
\end{table}

Table \ref{tables.ablation} shows the impact of global information on \textsc{Seg2Act}. We break down the global context stack formatted in the schema into two components: text and symbol. The symbol represents the hierarchical mark before the text, such as ``+'' and ``*''. Therefore, deleting the global context stack (symbol) means using only the texts in the schema, and deleting the global context stack (text) means using only symbols in the schema. We observe that:

1) \textbf{The structural representation schema offers an effective way to perceive the global document structure.} When hierarchical symbols are removed, \textsc{Seg2Act}'s ability to predict the overall document structure significantly diminishes, resulting in a decrease of 6.46 in DocAcc.

2) \textbf{There's an inherent trade-off between hierarchical prediction and paragraph concatenation with the use of the global context stack.} We notice a slight change of -0.06, -0.09, +0.3 in F1-score for total nodes, paragraph nodes and heading nodes, respectively, when hierarchical symbols are added. These symbols encourage \textsc{Seg2Act} to focus on hierarchical discrimination, slightly diminishing its ability to concatenate paragraphs and resulting in a minor decrease in F1-score.

\subsubsection{Effects of Multi-segment Multi-action}

\begin{table}[t]
\centering
\scalebox{0.64}{
    \begin{tabular}{c|ccccc}
    \toprule
    \multirow{2}{*}{\makecell{\textbf{Input} \\ \textbf{Segment Window}}} & \multicolumn{5}{c}{\textbf{Output Action Window}} \\
    \cline{2-6}
    & $w_{O}$ = 1 & $w_{O}$ = 2 & $w_{O}$ = 3 & $w_{O}$ = 4 & $w_{O}$ = 5 \\
    \midrule
    $w_I$ = 1 & \makecell{\textbf{91.56}\\(\textbf{10.43s})} & - & - & - & - \\
    \midrule
    $w_I$ = 2 & \makecell{\textbf{93.14}\\(17.61s)} & \makecell{92.99\\(\textbf{8.79s})} & - & - & - \\
    \midrule
    $w_I$ = 3 & \makecell{\textbf{92.83}\\(24.73s)} & \makecell{91.73\\(12.29s)} & \makecell{92.63\\(\textbf{8.13s})} & - & - \\
    \midrule
    $w_I$ = 4 & \makecell{\textbf{92.32}\\(31.23s)} & \makecell{91.41\\(15.83s)} & \makecell{91.76\\(10.48s)} & \makecell{91.74\\(\textbf{7.75s})} & - \\
    \midrule
    $w_I$ = 5 & \makecell{93.03\\(38.44s)} & \makecell{\textbf{93.06}\\(19.64s)} & \makecell{91.40\\(12.96s)} & \makecell{92.76\\(9.76s)} & \makecell{92.43\\(\textbf{7.63s})} \\
    \midrule
    Baseline & \multicolumn{5}{c}{89.55 (10.58s)} \\
    \bottomrule
    \end{tabular}
}
\caption{The F1-score of total nodes (inference time per document) of scaling the lengths of the input segment window and output action window for \textsc{Seg2Act} on ChCatExt. Baseline refers to TRACER in Baichuan-7B.}
\label{tables.scale_result}
\end{table}

To verify the effect of the multi-segment multi-action strategy on \textsc{Seg2Act}’s performance and efficiency, we scale the lengths of input segment window and output action window from 1 to 5, conducting experiments on ChCatExt. We take F1-score for total nodes as the metric and measure the average inference time for each document, as shown in Table \ref{tables.scale_result}. We can see that:

1) \textbf{Providing insights from the following consecutive segments mitigates short-sighted issue and enhances performance.}  Extending the input segment window length $w_I$ to 2, 3, 4, and 5, the \textsc{Seg2Act} method exhibits improvements in F1-score of +1.58, +1.27, +0.76, and +1.50, compared to the case where $w_I$ = 1.

2) \textbf{Simultaneously generating multiple actions ensures decoding efficiency of \textsc{Seg2Act}.}
By increasing the output action window length $w_O$, \textsc{Seg2Act} experiences a reduction in inference time while maintaining comparable performance. For instance, when comparing ($w_I=3$,$w_O=3$) with ($w_I=1$,$w_O=1$), \textsc{Seg2Act} demonstrates a notable improvement with a +1.07 increase in F1-score and a $\times 0.28$ boost in inference speed.

\begin{table*}[t]
\setlength{\belowcaptionskip}{-5pt}
\centering
\scalebox{0.58}{
    \begin{tabular}{lp{9.5cm}p{8cm}lc}
    \toprule
    \textbf{Method} & \textbf{Stack} & \textbf{Segment} & \textbf{Predicted Action} & \\
    \midrule
    \midrule
    TRACER$^*$ & \makecell[{{p{9.5cm}}}]{Risk Principle \textcolor{red}{$\hookleftarrow$}} & \makecell[{{p{8cm}}}]{Chapter 3 Basis and Scope for Determining the Holders of Employee Stock Ownership Plans \textcolor{red}{$\hookleftarrow$}} & New Paragraph & \ding{55} \\
    \midrule
    \makecell[l]{\textsc{Seg2Act}\\($w_I=w_O=1$)} & \makecell[{{p{9.5cm}}}]{
    + Summary of Employee Stock Ownership Plan (Draft) \textcolor{red}{$\hookleftarrow$}\\++ Chapter 2 Purpose and Basic Principles of Employee Stock Ownership Plans \textcolor{red}{$\hookleftarrow$}\\+++ 2. The basic principles of employee stock ownership plans \textcolor{red}{$\hookleftarrow$}\\++++ Risk Principle \textcolor{red}{$\hookleftarrow$}\\~* Participants in this employee stock ownership plan ... equal rights and interests with other investors. \textcolor{red}{$\hookleftarrow$}} & Chapter 3 Basis and Scope for Determining the Holders of Employee Stock Ownership Plans \textcolor{red}{$\hookleftarrow$} & New Level-2 Heading & \ding{51} \\
    \midrule
    \midrule
    \makecell[l]{\textsc{Seg2Act}\\($w_I=w_O=1$)} & \makecell[{{p{9.5cm}}}]{
    + Announcement on the Inquiry Letter on Matters Related to the Company's Application for Bankruptcy \textcolor{red}{$\hookleftarrow$}} & --- Is the early acquisition decision reasonable? \textcolor{red}{$\hookleftarrow$} & New Paragraph & \ding{55} \\
    \midrule
    \makecell[l]{\textsc{Seg2Act}\\($w_I=w_O=3$)} & \makecell[{{p{9.5cm}}}]{
    + Announcement on the Inquiry Letter on Matters Related to the Company's Application for Bankruptcy \textcolor{red}{$\hookleftarrow$}} & \makecell[{{p{8cm}}}]{--- Is the early acquisition decision reasonable? \textcolor{red}{$\hookleftarrow$}\\On January 20, 2021, the company announced that it would acquire 100\% equity of HNA Airport Group from its related party Hainan Airlines Travel Service Co., Ltd. for 500 million yuan, with a net asset value of 34.073 million euros. The transaction \textcolor{red}{$\hookleftarrow$}\\appreciation rate is about 87\%, and the main assets of HNA Airport Group are 82.5\% equity of Hahn Airport in Frankfurt, Germany (hereinafter referred to as Hahn Airport). In the short term, the company has announced that HNA Airport Group and Hahn Airport have filed for bankruptcy. \textcolor{red}{$\hookleftarrow$}} & \makecell[l]{New Level-2 Heading\\New Paragraph\\Concatenation} & \ding{51} \\
    \midrule
    \bottomrule
    \end{tabular}
}
\caption{A case study for models utilizing the Baichuan-7B backbone. }
\label{tables.case_study}
\end{table*}

\subsection{Analysis of Document Length}

To analyze the impact of document length, we show the performance on different subsets of ChCatExt in Figure \ref{figures.depth_and_token}. We can observe that: 

\begin{figure}[ht]
    \centering
    \begin{subfigure}[b]{0.4\textwidth}
        \centering
        \includegraphics[width=\textwidth]{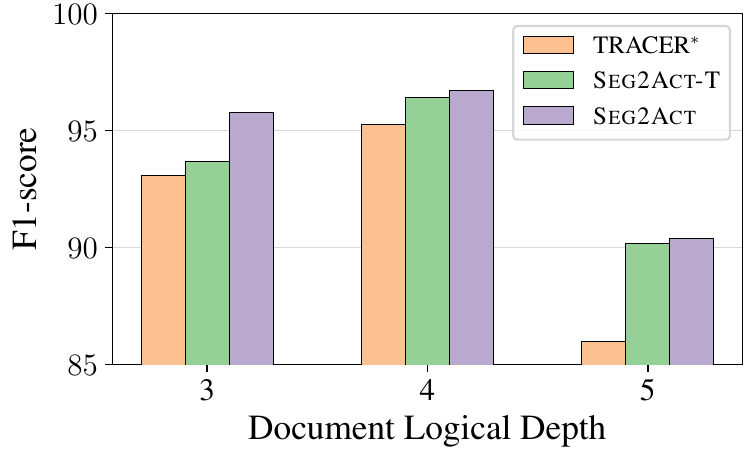}
        \caption{}
        \label{figures.depth}
    \end{subfigure}
    \hfill
    \begin{subfigure}[b]{0.4\textwidth}
        \centering
        \includegraphics[width=\textwidth]{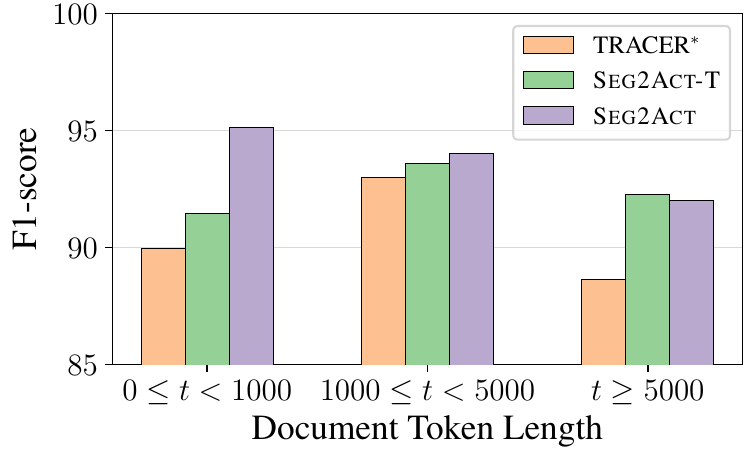}
        \caption{}
        \label{figures.token}
    \end{subfigure}
    \caption{Results (F1-score of total nodes) for documents with different logical tree depths (a) and token lengths (b) on ChCatExt dataset.}
    \label{figures.depth_and_token}
\end{figure}


1) \textbf{Our proposed actions are more effective for complex document logical structure than shift-reduce actions}. As the depth of the logical structure increases, the performance of all models significantly declines. However, \textsc{Seg2Act} still achieves the best performance among the three models.

2) \textbf{Global contextual information improves the logical structure handling of lengthy documents}. As document token length increases, models with global context experience a smaller performance drop compared to TRACER.

\subsection{Case Study}

We illustrate two cases in the prediction steps, as depicted in Table \ref{tables.case_study}. In the first scenario, the local pairwise method TRACER fails to predict the current input segment for the ``Reduce'' action due to a lack of global perspective. On the contrary, our \textsc{Seg2Act} successfully predicts the correct type and level with the assistance of the global context stack. In the second case, expanding the input segment window enables the model to make more insightful decisions. These two cases highlight the effectiveness of our method.

\section{Related Work}

Document logical structuring has received significant attention for an extended period \cite{DBLP:conf/icpr/TsujimotoA90,DBLP:phd/us/Summers98,DBLP:conf/drr/MaoRK03,DBLP:journals/ijdls/LuongNK10,DBLP:journals/nle/PembeG15,DBLP:conf/emnlp/GopinathWS18,maarouf-etal-2021-financial,DBLP:conf/icdar/ZhuZLYRWWHCC23}. Traditional methods have predominantly focused on designing heuristic or hand-crafted rules to extract logical structures \cite{fisher1991logical,DBLP:conf/icdar/Conway93}. For instance, text regular matching methods can be employed to differentiate headings from paragraphs. However, a notable drawback of such rule-based approaches is their specificity to certain document types, limiting their applicability to others.

In recent years, the advent of deep learning has opened up new avenues for document logical structuring, with a particularly promising trend being multi-modal and multi-stage modeling \cite{bourez-2021-fintoc,DBLP:journals/jcst/CaoCZL22}. From a multi-modal perspective, the incorporation of layout and vision modalities enhances the representation of semantic structures \cite{DBLP:conf/icpr/HuZZDW22,DBLP:conf/emnlp/Wang0023}. On the other hand, adopting a multi-stage approach involves decomposing the task into subtasks, which facilitates an easier and more manageable modeling process \cite{DBLP:conf/bdc/RahmanF17,DBLP:conf/icdar/BentabetJF19}. While multi-modal methods excel with single-page document images, they struggle to effectively model the intricate structures of lengthy, multi-page documents. Similarly, multi-stage methods encounter challenges related to error propagation when concatenating all stages in real-world applications.

Another noteworthy direction is the transition-based extraction \cite{DBLP:conf/acl-nllp/KoreedaM21,DBLP:conf/icdar/ZhuZLYRWWHCC23}. Transition-based methods parse texts into structured trees from the bottom up, offering efficiency and suitability for very long documents. However, these methods focus on pairwise local context, capturing only local information while neglecting the global information of the documents.

In contrast to previous works, our research introduces an end-to-end and generation-based method. This approach minimizes error propagation and enhances generalization. Furthermore, our framework, incorporating global context information, helps the action generation process and efficiently predicts the logical structure of documents.

\section{Conclusions}

This paper proposes \textsc{Seg2Act}, a novel method that models document logical structuring task as an end-to-end, one-pass action generation process. By leveraging the generative language model as an action generator and incorporating a global context stack, \textsc{Seg2Act} achieves significant performance and strong generalization on two benchmark datasets.
For future work, we plan to explore the integration of long-context language models and multi-modal language models with the \textsc{Seg2Act} framework.

\clearpage

\section*{Limitations}

First, generating indefinite-length action sequence using generative model may result in some cases that are challenging to parse, despite being constrained by hard rules. For example, in the multi-segment multi-action strategy, it cannot be guaranteed that the model will always output action sequence matching the specified $w_I$ count.

Second, our approach does not utilize visual information, thus requiring a proper order of input text segments, making it difficult to handle sequence with disrupted text segment order. Therefore, more effort is needed to incorporate visual information, making our method more flexible and applicable in a wider range of scenarios.

\section*{Ethics Statement}
In consideration of ethical concerns, we provide the following detailed descriptions:

1) All the data and backbone model weights we use come from publicly available sources. When using these resources for this study, we strictly adhere to their licensing agreements.

2) Our approach relies on large language models such as Baichuan-7B \cite{baichuan2023} as its backbone. As these language models have been trained on extensive text data sourced from the Web, it may be susceptible to issues such as toxic language and bias. However, our model is further fine-tuned to only generate structural actions and can only be used for document logical structuring, significantly mitigating the impact of these concerns.

\section*{Acknowledgements}

We sincerely thank the reviewers for their insightful comments and valuable suggestions. This work was supported by the Natural Science Foundation of China (No.  62306303, 62122077 and 62106251), and the Basic Research Program of ISCAS (Grant No. ISCAS-ZD-202402).

\bibliography{custom}

\clearpage
\appendix

\section{Data Pre-processing}
\label{data_preprocessing}

Currently, most datasets of document logical structuring are labeled with logical tree structure. In order to train our model, we convert the logical tree structure to our training corpus using preorder traversal, as illustrated in Algorithm \ref{algorithm.data}. 

\begin{algorithm}[!h]
\small
\SetAlgoLined
\SetKwInOut{Input}{Input}
\SetKwInOut{Output}{Output}
\SetKwInOut{Initialize}{Initialize}
\Input{Document logical tree structure $T$.}
\Output{Text segments $X=x_1,...,x_N$, \\action sequence $Y=y_1,...,y_N$.}
\Initialize{$X \gets$ [~], $Y \gets$ [~].}

\SetKwFunction{Travel}{Travel}
\SetKwProg{Fn}{Procedure}{:}{}
\Fn{\Travel{\textnormal{node}}}{
    $X$.extend( node.content )\;
    \uIf{\textnormal{node.type = ``Heading''}}{
        $Y$.append( ``+'' * len(node.depth) )\;
    }
    \Else{
        $Y$.append( ``*'' )\;
    }
    segment\_num $\gets$ len(node.content)\;
    \If{\textnormal{segment\_num > 1}}{
        \For{\textnormal{$i \gets $ 2 \KwTo segment\_num}}{
            $Y$.append( ``='' )\;
        }
    }
    child\_num $\gets$ len(node.children)\;
    \For{\textnormal{child $\in$ node.children}}{
        \Travel{\textnormal{child}}\;
    }
}
\Return $X,Y$ after \Travel{\textnormal{$T$.root}}\;

\caption{\nolbreaks{Logical structure to training data}}
\label{algorithm.data}
\end{algorithm}

\section{Action Constraints}
\label{constraints}

For \textsc{Seg2Act-T}, we conduct the same constraints as TRACER~\cite{DBLP:conf/icdar/ZhuZLYRWWHCC23}, which includes four actions: \textit{Sub-Heading}, \textit{Sub-Text}, \textit{Reduce}, \textit{Concat}. The constraints are as follows: 

\begin{itemize}
\setlength{\itemsep}{1pt}
    \item The action between Root node and the first input text segment can only be \textit{Sub-Heading} or \textit{Sub-Paragraph};
    \item The paragraph nodes can only be leaf nodes in the logical tree structure. Thus, if the last segment is predicted to be a paragraph node, only \textit{Reduce} and \textit{Concat} actions are permitted for the prediction of current segment.
\end{itemize}

For \textsc{Seg2Act}, the constraints are as follows:

\begin{itemize}
\setlength{\itemsep}{1pt}
    \item The predicted token must be in the predefined action set. We only allow token prediction in predefined set \{``+'', ``*'', ``='', ``$\backslash$n''\} and ban all other predictions through \verb|LogitsProcessor| of the Transformer library \cite{wolf2020transformers}, which supports forcibly setting token prediction probability to 0; 
    \item The \textit{Concatenation} action cannot be performed when the stack contains only the root node. Therefore, the action for the first input text segment can only be \textit{New Level-1 Heading} or \textit{New Paragraph} (indicating that the initial predicted token can only be ``+'' or ``*''). We also utilize \verb|LogitsProcessor| to execute this constraint;
    \item Heading nodes are prohibited from skipping levels, and if they do so, they are constrained to be at the current maximum level plus 1 (for example, if the generated action is ``++++'' but the maximum level of the heading nodes in the global context stack is only 2, we modify the decoded action to be \textit{New Level-3 Heading}). This constraint ensures that the parent node for newly added nodes can be found within the stack and the tree structure.    
\end{itemize}

For the first constraint of \textsc{Seg2Act}, different models may use different tokenizers, resulting in different token prediction strategies.
In addition, the tokens allowed to be predicted are also related to the model's last generated tokens.
Table \ref{tables.tokenizer} shows the allowed token predictions for the GPT2-Medium and Baichuan-7B models, respectively.

\begin{table}[h]
\centering
\begin{subtable}{0.48\textwidth}
\centering
\scalebox{0.7}{
    \begin{tabular}{c|p{2em}<{\centering}|p{2em}<{\centering}|p{2em}<{\centering}|p{2em}<{\centering}|p{2em}<{\centering}|p{2em}<{\centering}|p{2em}<{\centering}}
    \toprule
    \multirow{2}{*}{\textbf{Last Token}} & \multicolumn{6}{c}{\textbf{Next Token}} \\
     & $\backslash \texttt{n}$ & + & ++ & ++++ & * & = & $\texttt{</s>}$ \\
     \midrule
     $\backslash \texttt{n}$ & & $\surd$ & $\surd$ & $\surd$ & $\surd$ & $\surd$ & $\surd$ \\
     \midrule
     + & $\surd$ & $\surd$ & $\surd$ & $\surd$ & & & \\
     \midrule
     ++ & $\surd$ & $\surd$ & $\surd$ & $\surd$ & & & \\
     \midrule
     ++++ & $\surd$ & $\surd$ & $\surd$ & $\surd$ & & & \\
     \midrule
     * & $\surd$ & & & & & & \\
     \midrule
     = & $\surd$ & & & & & & \\
    \bottomrule
    \end{tabular}
}
\caption{The allowed token predictions in GPT2-Medium model.}
\label{tables.gpt_tokenizer}
\end{subtable}

\begin{subtable}{0.48\textwidth}
\centering
\scalebox{0.7}{
    \begin{tabular}{c|p{2em}<{\centering}|p{2em}<{\centering}|p{2em}<{\centering}|p{2em}<{\centering}|p{2em}<{\centering}|p{2em}<{\centering}}
    \toprule
    \multirow{2}{*}{\textbf{Last Token}} & \multicolumn{6}{c}{\textbf{Next Token}} \\
     & $\backslash \texttt{n}$ & + & ++ & * & = & $\texttt{</s>}$ \\
     \midrule
     $\backslash \texttt{n}$ & & $\surd$ & $\surd$ & $\surd$ & $\surd$ & $\surd$ \\
     \midrule
     + & $\surd$ & $\surd$ & $\surd$ & & & \\
     \midrule
     ++ & $\surd$ & $\surd$ & $\surd$ & & & \\
     \midrule
     * & $\surd$ & & & & & \\
     \midrule
     = & $\surd$ & & & & & \\
    \bottomrule
    \end{tabular}
}
\caption{The allowed token predictions in Baichuan-7B model.}
\label{tables.baichuan_tokenizer}
\end{subtable}
\setlength{\abovecaptionskip}{-5pt}
\setlength{\belowcaptionskip}{-5pt}
\caption{The allowed token predictions for models with different tokenizers.}
\label{tables.tokenizer}
\end{table}






\section{Effects of Model Size}

In this section, we explore the impact of model size on our proposed framework.

As demonstrated in Table \ref{tables.scaling_up}, enlarging models can boost performance, and among models of equal size, those integrating global context typically exhibit superior performance.

\begin{table}[t]
\centering
\scalebox{0.72}{
    \begin{tabular}{lccccc}
    \toprule
    \textbf{Method} & \textbf{Heading} & \textbf{Paragraph} & \textbf{Total} & \textbf{DocAcc} \\
    \midrule
    \multicolumn{5}{l}{\textit{\textbf{Methods using Baichuan-7B as Backbone}}} \\
    TRACER$^{*}$ & 94.91 & 91.62 & 89.55 & 53.85 \\
    \textsc{Seg2Act-T} (Ours) & 96.01 & 93.98 & 92.39 & 58.46 \\
    \textsc{Seg2Act} (Ours) & 96.01 & 94.19 & 92.63 & 63.69 \\
    \midrule
    \multicolumn{5}{l}{\textit{\textbf{Methods using Baichuan-13B as Backbone}}} \\
    TRACER$^{*}$ & 94.79 & 92.49 & 90.39 & 54.15 \\
    \textsc{Seg2Act-T} (Ours) & 95.97 & 93.73 & 92.06 & 60.62 \\
    \textsc{Seg2Act} (Ours) & \textbf{96.25} & \textbf{94.40} & \textbf{92.83} & \textbf{67.08} \\
    \bottomrule
    \end{tabular}
}
\caption{The result on ChCatExt (Heading, Paragraph, Total nodes in F1-score and logical structure accuracy at the document level).}
\label{tables.scaling_up}
\end{table}

\begin{table}[t]
\centering
\scalebox{0.8}{
    \begin{tabular}{l|cc|c}
    \toprule
    \textbf{Model} & \textbf{Total} & \textbf{DocAcc} & \textbf{TimeCost} \\
    \midrule
    Qwen1.5-0.5B & 92.22 & 57.54 & \textbf{4.01s} \\
    Qwen1.5-1.8B & \textbf{92.99} & 63.69 & 4.27s \\
    Qwen1.5-4B & 92.93 & \textbf{65.23} & 7.06s \\
    \midrule
    Baichuan-7B & 92.63 & 63.69 & 8.13s \\
    \bottomrule
    \end{tabular}
}
\caption{The result of \textsc{Seg2Act} on ChCatExt (Total nodes in F1-score, logical structure accuracy at the document level and time cost per document). }
\label{tables.scaling_down}
\end{table}

However, the performance gains from increasing the model size are not cost-effective compared to the expenses of training larger models. Additionally, larger models result in longer inference times, making efficiency a critical concern in practical applications. Therefore, we also discuss the performance of our proposed \textsc{Seg2Act} framework when decreasing the model size. 
Since there is no version of the Baichuan model smaller than 7B size, we choose Qwen1.5 model \cite{DBLP:journals/corr/abs-2309-16609} for experiments.
As shown in Table \ref{tables.scaling_down}, we can observe that:

1) \textbf{Backbone model choice affects performance}. Comparing the Qwen1.5 and Baichuan backbone models, the Qwen1.5-4B outperforms the Baichuan-7B in F1-score and document-level accuracy, while also being smaller in model size.

2) \textbf{The action generation framework may not necessarily require an oversize model}. For instance, in the Qwen1.5 series of models, the Qwen1.5-1.8B model achieves similar performance to the Qwen1.5-4B, but is $\times 0.65$ faster.

\end{document}